\documentclass[a4paper]{article}

\usepackage{times}
\usepackage{graphicx} 
\usepackage{subfigure} 
\usepackage{rotating}
\usepackage[square,sort,comma,numbers,compress]{natbib}

\usepackage[font=small,labelfont=bf]{caption}
\usepackage{algorithm}
\usepackage{algorithmic}
\usepackage{dsfont}
\usepackage{amsmath}
\usepackage{amsfonts}
\usepackage{amssymb}
\usepackage{authblk}

\usepackage{hyperref}

\title{Deep Networks with Internal Selective Attention through Feedback Connections \\ \small{\vspace{.4cm} \it A version of this paper was submitted to ICML 2014 on 31-01-2014.}}

\author[*]{Marijn Stollenga \texttt{marijn@idsia.ch}}
\author[*]{\authorcr Jonathan Masci \texttt{jonathan@idsia.ch}}
\affil[*]{Shared first author.}
\author[ ]{\authorcr Faustino Gomez \texttt{tino@idsia.ch}}
\author[ ]{\authorcr Juergen Schmidhuber \texttt{juergen@idsia.ch}}

\usepackage{xcolor}

\begin{document} 
\maketitle



\begin{abstract} 
  Traditional convolutional neural networks (CNN) are stationary and
  feedforward.  They neither change their parameters during evaluation
  nor use feedback from higher to lower layers. Real brains, however,
  do. So does our Deep Attention Selective Network (dasNet)
  architecture. DasNet’s feedback structure can dynamically alter its
  convolutional filter sensitivities during classification. It
  harnesses the power of sequential processing to improve
  classification performance, by allowing the network to iteratively
  focus its internal attention on some of its convolutional
  filters. Feedback is trained through direct policy search in a huge
  million-dimensional parameter space, through scalable natural
  evolution strategies (SNES). On the CIFAR-10 and CIFAR-100 datasets,
  dasNet outperforms the previous state-of-the-art model.
\end{abstract}

\section{Introduction}
\label{sec:introduction}

Deep convolutional neural networks (CNNs) \cite{Fukushima:1979neocognitron} with max-pooling layers \cite{weng1992} trained by backprop \cite{LeCun:89,ranzato:2007,scherer:2010} on GPUs \cite{ciresan:2011ijcai} have become
the state-of-the-art in object recognition~\cite{ciresan2012cvpr,Krizhevsky:2012,wan2013regularization,goodfellow2013maxout},
segmentation/detection~\cite{miccai2013,Ciresan:2012f},
and scene parsing~\cite{farabet2013learning,sermanet-cvpr-2013,sermanet2013overfeat} (for an extensive review 
see \cite{schmidhuber2014deep}).
These architectures consist of many stacked feedforward layers,
mimicking the bottom-up path of the human visual cortex, where each
layer learns progressively more abstract representations of the input
data. Low-level stages tend to learn biologically plausible feature
detectors, such as Gabor filters~\citep{gabor1946}.  Detectors in higher layers learn to
respond to concrete visual objects or their
parts, e.g., \cite{zeiler2013visualize,simonyan2013visual,Zeiler2011AdaptiveDeconvolutionalNetworks,quoc:2012}.
Once trained, the CNN never changes its weights or filters during
evaluation.


Evolution has discovered efficient feedforward pathways for 
recognizing certain objects in the blink of an eye.  However, an
expert ornithologist, asked to classify a bird belonging to one of two
very similar species, may have to think for more than a few
milliseconds before answering~\cite{branson2010visual,
  WelinderEtal2010}, implying that several feedforward evaluations are
performed, where each evaluation tries to elicit different information
from the image.  Since humans benefit greatly from this strategy, we
hypothesize CNNs can too.  This requires: (1) the formulation of a
non-stationary CNN that can adapt its own behaviour post-training, and
(2) a process that decides \emph{how} to adapt the CNNs behaviour.

This paper introduces Deep Attention Selective Networks (dasNet) which
model selective attention in deep CNNs by allowing each layer to
influence all other layers on successive passes over an image through
special connections (both bottom-up and top-down), that modulate the
activity of the convolutional filters.  The weights of these special
connections implement a control policy that is learned through
reinforcement learning {\em after} the CNN has been trained in the
usual way via supervised learning. Given an input image, the
attentional policy can enhance or suppress 
features over multiple passes to improve the classification of
difficult cases not captured by the initially supervised training.
Our aim is to let the system check the usefulness of internal CNN filters 
automatically, omitting manual inspection~\cite{zeiler2013}.

In our current implementation, the attentional policy is evolved using
Separable Natural Evolution Strategies (SNES; ~\cite{schaul2011high}),
instead of a conventional, single agent reinforcement learning method
(e.g. value iteration, temporal difference, policy gradients, etc.)
due to the large number of parameters (over 1 million)
required to control CNNs of the size typically used in image
classification.  Experiments on
CIFAR-10 and CIFAR100~\cite{krizhevsky:2009} 
show that on difficult classification instances, the network corrects
itself by emphasizing and de-emphasizing certain filters,
outperforming a previous state-of-the-art CNN.

\section{Maxout Networks}
\label{sec:maxout}
In this work we use the Maxout networks~\cite{goodfellow2013maxout},
combined with dropout~\cite{hinton2012improving}, as the underlying
model for dasNet. Maxout networks represent the state-of-the-art for
object recognition in various tasks and have only been outperformed
(by a small margin) by averaging committees of several convolutional
neural networks.  A similar approach, which does not reduce
dimensionality in favor of sparsity in the representation has also
been recently presented~\cite{srivastava:2013}.  Maxout CNNs consist
of a stack of alternating convolutional and maxout layers, with a
final classification layer on top:

\paragraph{Convolutional Layer.} The input to this layer can be an
image or the output of a previous layer, consisting of $c$ input maps
of width $m$ and height $n$: $x \in \mathbb{R}^{c \times m \times n}$.
The output consists of a set of $c'$ output maps: $y \in
\mathbb{R}^{c' \times m' \times
  n'}$. 
The convolutional layer is parameterized by $c \cdot c'$ filters of size $k \times k$.
We denote the filters by $F^\ell_{i, j} \in \mathbb{R}^{k \times k}$, where $i$ and $j$ are indexes of the input and output maps and
$\ell$ denotes the layer.
\begin{align}
\label{eq:conv}
y_j^\ell = \sum_{i=0}^{i=c}\phi (x_i \ast F_{i,j}^\ell)
\end{align}
\noindent
where $i$ and $j$ index the input and output map respectively, $\ast$
is the convolutional operator, $\phi$ is an element-wise nonlinear
function, and $\ell$ is used to index the layer.  The size of the
output is determined by the kernel size and the stride used for the
convolution (see \cite{goodfellow2013maxout}).



\paragraph{Pooling Layer.} A pooling layer is used to reduced the
dimensionality of the output from a convolutional layer. The usual
approach is to take the maximum value among non-~or
partially-overlapping patches in every map, therefore reducing
dimensionality along the height and width \cite{weng1992}. Instead, a Maxout
pooling layer reduces every $b$ consecutive maps to one map, by keeping only 
the maximum value for every pixel-position, where $b$ is
called the block size. Thus the map reduces $c$ input maps to $c' = c / b$ output maps.
\begin{align}
\label{eq:pool}
y_{j,x,y}^{\ell} = \max_{i=0}^b y^{\ell-1}_{j\cdot b+i,x,y}
\end{align}
\noindent
where $y^\ell \in \mathbb{R}^{c' \times m' \times n'}$, and $\ell$ again is used to index the
layer.  The output of the pooling layer can either be used as input to
another pair of convolutional- and pooling layers, or form input to a
final classification layer.

\paragraph{Classification Layer.}
Finally, a classification step is performed. First the output of the last pooling
layer is flattened into one large vector $\vec{x}$, to form the input to the following equations:
\begin{align}
\bar{y}_j^\ell = \max_{i=0..b}F_{j\cdot b+i}^\ell \vec{x} \label{eq:maxclass}\\
\mathbf{v} = \sigma(F^{\ell+1} \bar{y}^\ell)\label{eq:softmax1}
\end{align}
where $F^\ell \in
\mathds{R}^{N \times |\vec{x}|}$ ($N$ is chosen), and $\sigma(\cdot)$
is the softmax activation function which produces the class
probabilities $\mathbf{v}$. The input is projected
by $F$ and then reduced using a maxout, similar to the pooling layer
(\ref{eq:maxclass}).  





\section{Reinforcement Learning}
\label{sec:rl}

Reinforcement learning (RL) is a general framework for learning to
make sequential decisions order to maximize an external reward
signal~\cite{Kaelbling1996, Sutton1998}.  The learning agent can be anything
that has the ability to \emph{act} and \emph{perceive} in a given
environment.

At time $t$, the agent receives an observation $o_t \in O$ of the
current state of the environment $s_t \in S$, and selects an action,
$a_t \in A$, chosen by a policy $\pi: O \to A$, where $S, O$ and $A$
the spaces of all possible states, observations, and action,
respectively.\footnote{In this work $\pi: O \to A$ is a deterministic
  policy; given an observation it will always output the same
  action. However, $\pi$ could be extended to stochastic policies.}
The agent then enters state $s_{t+1}$ and receives a reward $r_t \in
\mathds{R}$.  The objective is to find the policy, $\pi$, that
maximizes the expected future discounted reward,
$E[\sum_t\gamma^tr_t]$, where $\gamma\in[0,1]$ discounts the future,
modeling the ``farsightedness'' of the agent.

In dasNet, both the
observation and action spaces are real valued $O = \mathds{R}^{dim(O)}$,
$A = \mathds{R}^{dim(A)}$.  Therefore,  policy $\pi_{\theta}$ must  be represented by a function
approximator, e.g. a neural network, parameterized by $\theta$. 
Because the policies used to control the attention of the dasNet have
state and actions spaces of close to a thousand dimensions, the policy
parameter vector, $\theta$, will contain close to a million weights,
which is impractical for standard RL methods.
Therefore, we instead evolve the policy using a variant for Natural
Evolution Strategies (NES; \cite{wierstra2008natural, glasmachers2010exponential}), 
called Separable NES (SNES; \cite{schaul2011high}).
The NES family of black-box optimization algorithms use parameterized
probability distributions over the search space, instead of an
explicit population (i.e., a conventional ES \cite{Rechenberg:71,Schwefel:74,Holland:75}).
Typically, the distribution is a multivariate Gaussian parameterized by mean $\mu$
and covariance matrix $\Sigma$.  Each epoch a generation is sampled
from the distribution, which is then updated the direction of the
natural gradient of the expected fitness of the distribution.  SNES
differs from standard NES in that instead of maintaining the full
covariance matrix of the search distribution, uses only the diagonal
entries.  SNES is theoretically less powerful than standard NES, but
is substantially more efficient.



\section{Deep Attention Selective Networks (dasNet)}
\label{sec:model}

The idea behind dasNet is to harness the power of sequential
processing to improve classification performance by allowing the
network to iteratively focus the attention of its filters.  First, the
standard Maxout net (see Section~\ref{sec:maxout}) is augmented to
allow the filters to be weighted differently on different passes over
the same image (compare to equation~\ref{eq:conv}):
\begin{align}
\label{eq:gconv}
y_j^\ell = a_j^\ell\sum_{i=0}^{i=c}\phi (x_i \ast F_{i,j}^\ell),
\end{align}
\noindent
where $a_{j}^\ell$ is the weight of the $j$-th output map in layer
$\ell$, changing the strength of its activation, \emph{before}
applying the maxout pooling operator.  The vector ${\mathbf a}=[a^0_0,
a^0_1, \cdots, a^0_{c'},\\ a^1_0, \cdots, a^1_{c'}, \cdots]$ represents
the action that the learned policy must select in order to
sequentially focus the attention of the Maxout net on the most
discriminative features in the image being processed.
Changing action $\mathbf{a}$ will alter the behaviour of the CNN,
resulting in different outputs, even when the image $x$ does not
change.  We indicate this with the following notation:
\begin{equation}
\mathbf{v}_t = \mathbf{M}_t(\theta, x)
\end{equation}
\noindent where $\theta$ is the parameter vector of the policy,
$\pi_\theta$, and $\mathbf{v}_t$ is the output of the network on pass $t$.

\begin{algorithm}[t]                      
\caption{{\sc Train dasNet} ($\mathbf{M}$, $\mu$, $\Sigma$, $p$, $n$)}          
\label{alg1}                           
\begin{algorithmic}[1]                    
\WHILE{True}
\STATE $images \Leftarrow$ {\sc NextBatch}($n$)
    \FOR{$i = 0 \to p$}
        \STATE $\theta_i \sim \mathbb{N}(\mu, \Sigma)$
        \FOR{$j = 0 \to n$}
             \STATE $\mathbf{a}_0 \Leftarrow \mathds{1}$ \COMMENT{Initialize gates $a$ with identity activation}
             \FOR{$t = 0 \to T$}
                   \STATE $\mathbf{v}_t = \mathbf{M}_t(\theta_i, x_i)$
                   \STATE $\mathbf{o}_t \Leftarrow h(\mathbf{M}_t)$
                   \STATE $\mathbf{a}_{t+1} \Leftarrow \pi_{\theta_i}(\mathbf{o}_t)$
              \ENDFOR
              \STATE $L_i = -\lambda_{\text{boost}} d \log(\mathbf{v}_T)$
         \ENDFOR
         \STATE $\mathcal{F}[i] \Leftarrow f(\theta_i)$
         \STATE $\Theta[i] \Leftarrow \theta_i$
    \ENDFOR
        \STATE {\sc UpdateSNES}($\mathcal{F}$, $\Theta$)
\ENDWHILE
\end{algorithmic}
\end{algorithm}

Algorithm \ref{alg1} describes the dasNet training algorithm.  Given a
Maxout net, $\mathbf{M}$, that has already been trained to classify
images using training set, X, the policy, $\pi$, is evolved using SNES
to focus the attention of $\mathbf{M}$.  Each pass through the {\tt
  while} loop represents one generation of SNES.  Each generation
starts by selecting a subset of $n$ images from X at random.
Then each of the $p$ samples drawn from the SNES search distribution
(with mean $\mu$ and covariance $\Sigma$) representing the parameters,
$\theta_i$, of a candidate policy, $\pi_{\theta_i}$, undergoes $n$ trials,
one for each image in the batch.  During a trial, the image is
presented to the Maxout net $T$ times.  In the first pass, $t=0$, the
action, ${\mathbf a}_0$, is set to $a_{i}=1, \forall i$, so that
the Maxout network functions as it would normally --- the action has no effect.
Once the image is propagated through the net, an observation vector,
${\mathbf o}_0$, is constructed by concatenating the following
values extracted from $\mathbf{M}$, by $h(\cdot)$:
\begin{enumerate}
\item the average activation of \emph{every} output map $Avg(y_j)$
  (Equation~\ref{eq:pool}), of each Maxout layer.
\item the intermediate activations $\bar{y}_j$ of the classification
  layer.
\item the class probability vector,  $\mathbf{v}_t$.
\end{enumerate}
 While 
averaging map activations provides only partial state information,
these values should still be meaningful enough to allow for the
selection of good actions.
The candidate policy then maps the observation to an action:
\begin{align}
\pi_{\theta_i}(\mathbf{o}) = dim(A) \sigma(\boldsymbol{\theta}_i
\mathbf{o_t}) = \mathbf{a_t},
\end{align}
where $\boldsymbol{\theta} \in \mathbb{R}^{dim(A) \times dim(O)}$ is
the weight matrix of the neural network, and $\sigma$ is the softmax.
Note that the softmax function is scaled by the dimensionality of the
action space so that elements in the action vector average to $1$
(instead of regular softmax which \emph{sums} to $1$), ensuring that
all network outputs are positive, thereby keeping the filter
activations stable.

\begin{figure*}[t]
\centering
\includegraphics[width=\linewidth]{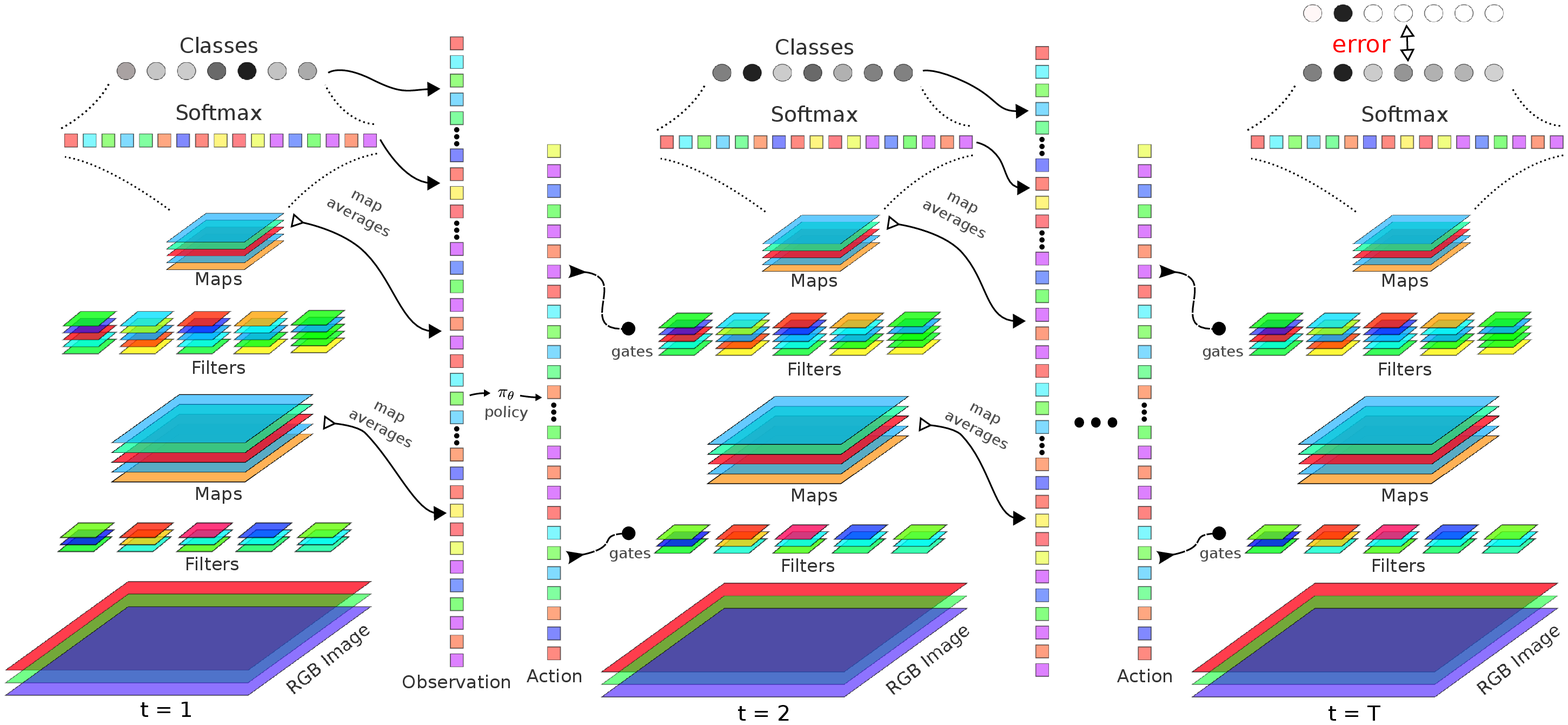}
\caption{{\bf The dasNet Network}. Each image in classified after $T$
  passes through the network.  After each forward propagation through
  the Maxout net, the output classification vector, the output of the
  second to last layer, and the averages of all feature maps, are
  combined into an observation vector that is used by a
  deterministic policy to choose an action that
  changes the weights of all the feature maps for the next pass of the
  same image.  After pass $T$, the output of the Maxout net is finally
used to classify the image.}
\label{fig:model}
\end{figure*}

On the next pass, the same image is processed again, but this time
using the filter weighting, ${\mathbf a}_1$.  This cycle is
repeated until pass $T$ (see figure~\ref{fig:model} for a
illustration of the process), at which time the performance of the
network is scored by:

\begin{align}
L_i = -\lambda_{\text{boost}} d \log(\mathbf{v}_T)\\
\mathbf{v}_T = \mathbf{M}_T(\theta_i, x_i)
\end{align}
\begin{align}
\lambda_{\text{boost}} = \begin{cases}
   \lambda_{\text{correct}}     & \text{if } d = \|\mathbf{v}_T\|_{\infty} \\
   \lambda_{\text{misclassified}}       & \text{otherwise,}
  \end{cases} 
\end{align}
\noindent
where $\mathbf{v}$ is the output of $\mathbf{M}$ at the end of the
pass $T$, $d$ is the correct classification, and $\lambda_{correct}$
and $\lambda_{misclassified}$ are constants.  $L_i$ measures the
weighted loss, where misclassified samples are weighted higher than
correctly classified samples $\lambda_{misclassified} >
\lambda_{correct}$.  This simple form of boosting is used to focus on
the `difficult' misclassified images.  Once all of the input images
have been processed, the policy is assigned the fitness:
\begin{align}
f(\theta_i) = \overbrace{\sum_{i=1}^n{L_i}}^{\text{cumulative score}} + \ \
\ \overbrace{\lambda_{L2} \| \theta_i \|_2}^{\text{regularization}}
\end{align}
\noindent
where $\lambda_{L2}$ is a regularization parameter.

Once all of the candidate policies have been evaluated, SNES updates
its distribution parameters ($\mu, \Sigma$) according the natural
gradient calculated from the sampled fitness values, $\mathcal{F}$.
As SNES repeatedly updates the distribution over the course of many
generations, the expected fitness of the distribution improves, until
the stopping criterion is met.




\section{Related Work}
Human vision is still the most advanced and flexible perceptual system
known.  Architecturally, visual cortex areas are highly connected,
including direct connections over multiple levels and top-down
connections. \citet{felleman1991distributed} constructed a (now
famous) hierarchy diagram of 32 different visual cortical areas in
macaque visual cortex.  About 40\% of all pairs of areas were
considered connected, and most connected areas were connected 
bidirectionally.  The top-down connections are more numerous than
bottom-up connections, and generally more
diffuse~\cite{douglas1995recurrent}.  They are thought to play 
primarily a modulatory role, while feedforward connections serve as
directed information carriers~\cite{Bullier2004}.

Analysis of response latencies to a newly-presented image lends
credence to the theory that there are two stages of visual processing:
a fast, pre-attentive phase, due to feedforward processing, followed
by an attentional phase, due to the influence of recurrent
processing~\cite{lamme2000distinct}.  After the feedforward pass, we
can recognize and localize simple salient stimuli, which can
``pop-out''~\cite{Itti:2007}, and response times do not increase
regardless of the number of distractors.  However, this effect has
only been conclusively shown for basic features such as color or
orientation; for categorical stimuli or faces, whether there is a
pop-out effect remains
controversial~\cite{francolini1979perceptual,vanrullen2006second}.
Regarding the attentional phase, feedback connections are known to
play important roles, such as in feature
grouping~\cite{gilbert2007brain}, in differentiating a foreground from
its background, (especially when the foreground is not highly
salient~\cite{hupe1998cortical,bullier2001role}), and perceptual
filling in~\cite{lamme2001blindsight}. Work by~\citet{bar2006top}
supports the idea that top-down projections from prefrontal cortex
play an important role in object recognition by quickly extracting
low-level spatial frequency information to provide an initial guess
about potential categories, forming a top-down expectation that biases
recognition.  Recurrent connections seem to rely heavily on
competitive inhibition and other feedback to make object recognition
more robust~\cite{wyatte2012, wyatte2012b}.

In the context of computer vision, RL has been shown to be able to
learn saccades in visual scenes to learn selective
attention~\cite{SchmidhuberHuber:91}, learn feedback to lower levels
\cite{oreilly1996, fukushima2003}, and improve face recognition
\cite{larochelle2010learning,goodrich2012reinforcement,stollenga2011using}.
It has been shown to be effective for object recognition
\cite{oreilly2013}, and has also been combined with traditional
computer vision primitives ~\cite{Whitehead:92}. Iterative processing
of images using recurrency has been successfully used for image reconstruction
\cite{behnke2001learning} and face-localization \cite{behnke2005face}.
All these approaches show that recurrency in processing and an RL perspective can lead to
novel algorithms that improve performance.  However, this research is
often applied to simplified datasets for demonstration purposes due to
computation constraints, and are not aimed at improving the
state-of-the-art.  In contrast, we apply this perspective directly to
the known state-of-the-art neural networks to show that this approach
is now feasible and actually increases performance.

\section{Experiments on CIFAR-10/100}
\label{sec:experiments}

The experimental evaluation of dasNet focuses on ambiguous
classification cases in the CIFAR-10 and CIFAR-100 data sets where,
due to a high number of common features, two classes are often
mistaken for each other.  These are the most interesting cases for our
approach.  By learning on top of an already trained model, dasNet must
aim at fixing these erroneous predictions without disrupting, or
forgetting, what has been learned.

The CIFAR-10 dataset~\cite{krizhevsky:2009} is composed of $32 \times
32$ color images split into $5 \times 10^4$ training and $10^4$
testing samples, where each image is assigned to one of $10$ classes.
The CIFAR-100 is similarly composed, but contains $100$ classes.



The number of steps, $T$, for the RL was experimentally determined and
fixed at $5$; enough steps to allow dasNet to adapt while being small
enough to be practical.  While it is be possible to iterate until some
condition is met, this could be a serious limitation in real-time
applications where predictable processing latency is critical.  In all
experiments we set $\lambda_{\text{correct}} = 0.005$,
$\lambda_{\text{misclassified}} = 1$ and $\lambda_{\text{L2}} =
0.005$.

\begin{figure}[t]
\begin{minipage}[b]{0.55\linewidth}
\begin{tabular}{l c c}
  \hline
  Method & {\footnotesize CIFAR-10} & {\footnotesize  CIFAR-100} \\
  \hline
  \hline
  Dropconnect~\cite{wan2013regularization}
& 9.32\% & -\\
  \hline
  Stochastic Pooling~\cite{2013:zeiler_stochpool} & 15.13\% & - \\ 
  Multi-column CNN~\cite{ciresan2012cvpr} & 11.21\% & - \\
  Maxout~\cite{goodfellow2013maxout} & 9.38\% & 38.57\% \\
  Maxout (our model) & 9.61\% & 34.54\% \\
  \bf{dasNet} & \bf{9.22}\%  & \bf{33.78}\% \\ 
  \hline
\end{tabular}
\captionof{table}{Classification results on CIFAR-10 and CIFAR-100
  datasets. The error on the test-set is shown for several methods.
  Note that the result for Dropconnect is the average of 12
  models. Our method improves over the state-of-the-art reference
  implementation to which feedback connections are added.  }
\label{tab:cifar10}
\end{minipage}
\quad
\begin{minipage}[b]{0.4\linewidth}
\centering
\includegraphics[angle=-90,width=\linewidth]{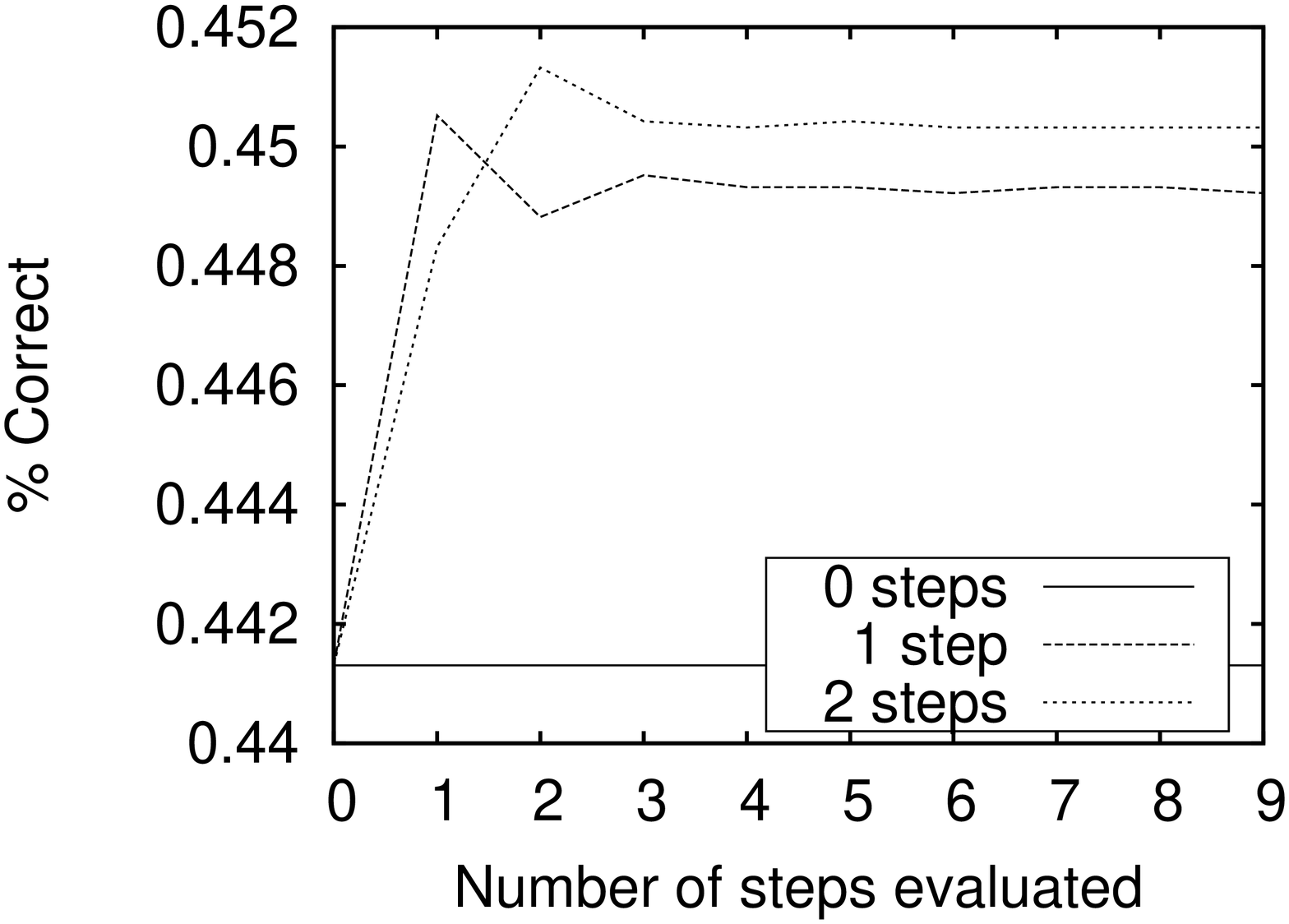}
\caption{Two dasNets were trained on CIFAR-100 for different values of $T$.
Then they were allowed to run for [0..9] iterations for each image. 
The performance peeks at the number of steps that the network is trained on, 
after which the performance drops, but does not explode, showing the dynamics are stable.}
\label{fig:1step2step}
\end{minipage}
\hspace{0.5cm}
\end{figure}


The Maxout network, $\mathbf{M}$, used in the experiments was trained
with data augmentation following the suggested global contrast
normalization and ZCA normalization protocol.  The model consists of
three convolutional maxout layers followed by a fully connected maxout
and softmax outputs. Dropout of $0.5$ was used in all layers except the
input layer, and $0.2$ for the input layer.  The population size for
SNES was set to 50.


Table~\ref{tab:cifar10} shows the performance of dasNet vs. other
methods, where it achieves a relative improvement of $6\%$ with
respect to the vanilla CNN.  This establishes a new state-of-the-art
result for this challenging dataset.

\begin{figure*}[t]
\centering
\includegraphics[width=\linewidth]{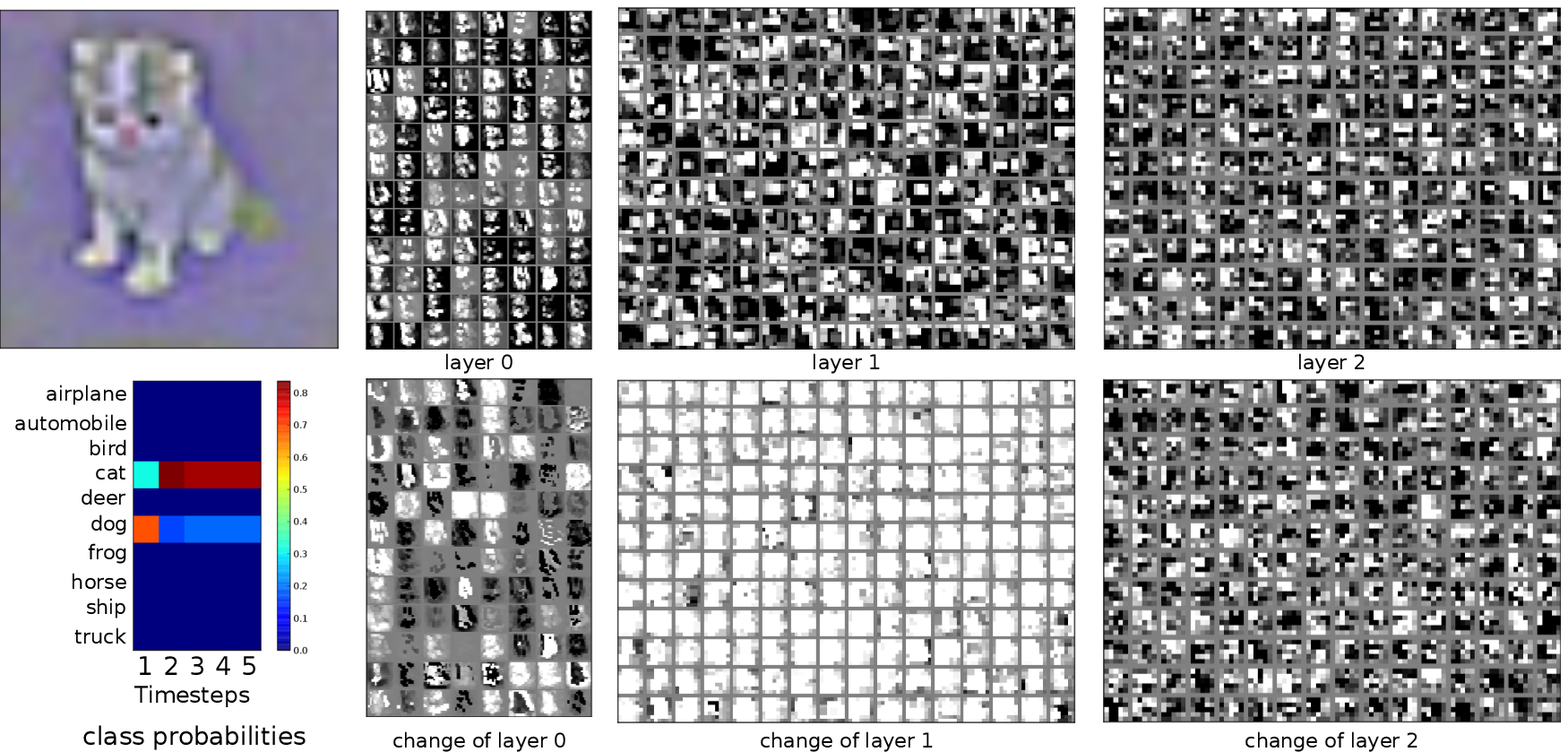}
\caption{The classification of a cat by the dasNet is shown. All
  output map activations in the final step are shown on the top. Their
  changes relative to initial activations in the first step
  are shown at the bottom (white = emphasis, black = suppression). The
  changes are normalized to show the effects more
  clearly. The class probabilities over time are shown on the left. The
  network first classifies the image as a dog (wrong) but corrects
  itself by emphasizing its convolutional filters to see it is actually a cat.}
\label{fig:details1}
\end{figure*}

Figure~\ref{fig:details1} shows the classification of a cat-image from
the test-set.  All output map activations in the final step are shown
at the top. The difference in activations compared to the first step,
i.e., the (de-)emphasis of each map, is shown on the bottom. On the
left are the class probabilities for each time-step. At the first
step, the classification is `dog', and the cat could indeed be mistaken
for a puppy. Note that in the first step, the network has not yet
received any feedback.  In the next step, the probability for `cat' goes up
dramatically, beating 'dog', and subsequently drops a bit in the
following steps.  The network has successfully disambiguated a cat
from a dog. If we investigate the filters, we see that already in the
lower layer emphasis changes significantly. Some filters focus more on
surroundings whilst others de-emphasize the eyes. 

In the second layer, almost all output maps are
emphasized. In the third and highest convolutional layer, the most
complex changes to the network. At this level the
positional correspondence is largely lost, and the filters are known
to code for `higher level' features. It is in this layer that changes
are the most influential because they are closest to the final output
layers. 
It is hard to analyze the effect of the alterations, but we
can see that the differences are not simple increases or decreases of
the output maps, as we then would expect the final activations and
their corresponding increases to be largely similar. Instead we see
complex emphasis and pattern suppression.

\paragraph{Dynamics} To investigate the dynamics, a small 2-layer dasNet
network was trained for different values of $T$. Then they were evaluated
by allowing them to run for $[0..9]$ steps.
Figure~\ref{fig:1step2step} shows results of training dasNet on CIFAR-100 for $T=1$ and $T=2$.
The performance goes up from the vanilla CNN, peaks at the $step = T$ as
expected, and reduces but stays stable after that.
So even though the dasNet was trained using only a small number of steps,
the dynamics stay stable when these are evaluated for as many as 10 steps.

To verify whether the dasNet policy is actually making good use of its gates,
their information content is estimated the following way:
The gate values in the last step are taking and used directly for classification.
If the gates are used properly then their activation should contain information that is relevant for classification
and we would expect a

dasNet that was trained with $T=2$ and are used as
features for classification. Then using \emph{only} the final gate-values (so without e.g. the output of the classification layer), 
a classification using 15-nearest neighbour and logistic regression was performed. This
resulted in a performance of \emph{40.70\%} and \emph{45.74\%} correct
respectively, similar to the performance of dasNet, confirming that
they contain significant information and we can conclude that they are purposefully used.

\section{Conclusion}
DasNet is a deep neural network with feedback connections that are
learned by through reinforcement learning to direct selective internal
attention to certain features extracted from images. After a rapid
first shot image classification through a standard stack of
feedforward filters, the feedback can actively alter the importance of
certain filters ``in hindsight'', correcting the initial guess via
additional internal ``thoughts''.

DasNet successfully learned to correct image misclassifications
produced by a fully trained feedforward Maxout network. Its active,
selective, internal spotlight of attention enabled state-of-the-art
results.

Future research will also consider more complex actions that spatially
focus on (or alter) parts of observed images.

\section*{Acknowledgments} 
We acknowledge Matthew Luciw, who provided a short literature review, partially included in the Related Work section.

{
\small
\bibliography{biblio,bib}
\bibliographystyle{unsrtnat}
}

\end{document}